%% file: main.tex
\documentclass[10pt,twocolumn,letterpaper]{article}

\usepackage[pagenumbers]{cvpr} 

\input{preamble}

\definecolor{cvprblue}{rgb}{0.21,0.49,0.74}
\usepackage[pagebackref,breaklinks,colorlinks,citecolor=cvprblue]{hyperref}

\usepackage{multirow}
\usepackage{makecell}
\usepackage{colortbl}
\usepackage{bm}

\definecolor{Last}{RGB}{240,204,126}
\definecolor{Transfer}{RGB}{191,191,225}
\definecolor{Avg}{RGB}{225, 191, 192}
\definecolor{Forgetting}{RGB}{255, 137, 137}

\title{CoLeCLIP: Open-Domain Continual Learning via Joint Task Prompt and Vocabulary Learning}

\author{Yukun Li$^{1}$,
Guansong Pang$^2$,
Wei Suo$^1$,
Chenchen Jing$^3$,
Yuling Xi$^1$, \\
Lingqiao Liu$^4$,
Hao Chen$^3$,
Guoqiang Liang$^{1}$,
Peng Wang$^{1}$ \\
$^1$School of Computer Science and Ningbo Institute, Northwestern Polytechnical University, Xi'an, China\\
$^2$School of Computing and Information Systems, Singapore Management University, Singapore \\
$^3$School of Computer Science, Zhejiang University, Hangzhou, China\\
$^4$School of Computer Science, University of Adelaide, Adelaide, Australia\\
}

\begin{document}
\maketitle
\input{sec/0_abstract}    
\input{sec/1_intro}
\input{sec/2_related_work}
\input{sec/3_method}

\input{sec/4_experiment}

\input{sec/5_conclusion}
{
    \small
    \bibliographystyle{ieeenat_fullname}
    \bibliography{main}
}

\input{sec/X_suppl}

\end{document}

%% file: preamble.tex
\usepackage[dvipsnames]{xcolor}


%% file: sec/0_abstract.tex
\begin{abstract}
This paper explores the problem of continual learning (CL) of vision-language models (VLMs) in open domains, where the models need to perform continual updating and inference on a streaming of datasets from diverse seen and unseen domains with novel classes. Such a capability is crucial for various applications in open environments, \eg, AI assistants, autonomous driving systems, and robotics. Current CL studies mostly focus on closed-set scenarios in a single domain with known classes. Large pre-trained VLMs like CLIP have demonstrated superior zero-shot recognition ability, and a number of recent studies leverage this ability to mitigate catastrophic forgetting in CL, but they focus on closed-set CL in a single domain dataset. Open-domain CL of large VLMs is significantly more challenging due to 1) large class correlations and domain gaps across the datasets and 2) the forgetting of zero-shot knowledge in the pre-trained VLMs in addition to the knowledge learned from the newly adapted datasets. In this work we introduce a novel approach, termed CoLeCLIP, that learns an open-domain CL model based on CLIP. It addresses these challenges by a joint learning of a set of task prompts and a cross-domain class vocabulary. Extensive experiments on 11 domain datasets show that CoLeCLIP outperforms state-of-the-art methods for open-domain CL under both task- and class-incremental learning settings.
\end{abstract}

%% file: sec/1_intro.tex
\section{Introduction}
\label{sec:intro}

\begin{figure}
    \centering
    \includegraphics[width=1\linewidth]{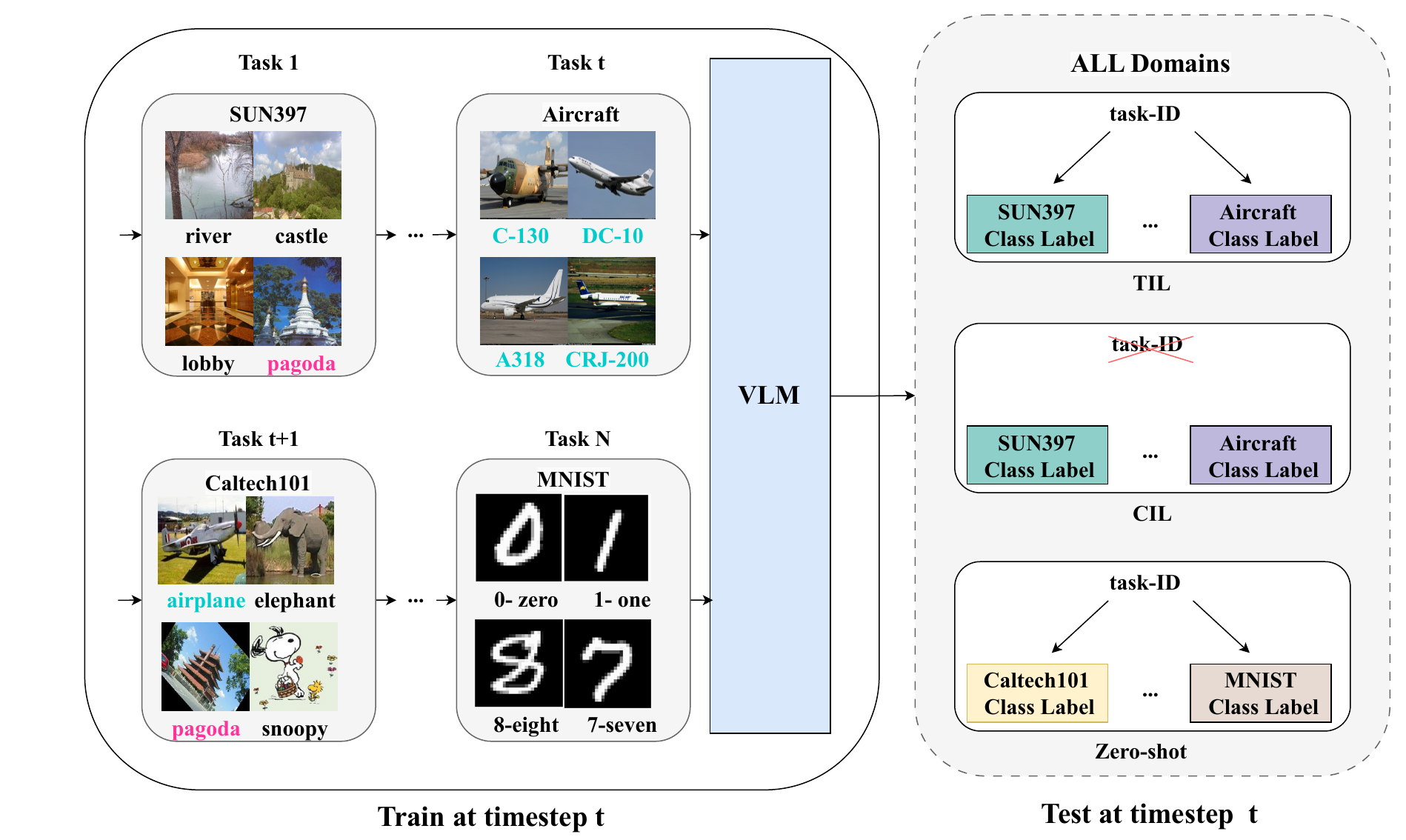}
    \caption{ 
    Illustration of ODCL, in which VLMs continuously acquire new knowledge from different datasets in incoming tasks, where different tasks may exhibit large domain gaps and strong class correlations. After training on the $t$-th task, VLM is evaluated on the datasets from all domains under task-incremental learning (TIL) or class-incremental learning (CIL).} 
    \label{fig:graph1}
\end{figure}

Vision-Language Models (VLMs) \cite{radford2021learning,jia2021scaling,li2023blip} learn visual concepts from natural language supervision and construct a joint multi-modal embedding space. A number of recently emerging VLMs, such as CLIP \cite{radford2021learning}, have shown remarkable zero-shot capabilities after pre-training on a vast amount of paired image-text data. However, they still struggle to perform well on certain domains that do not have sufficient data in their pre-training data.

One popular solution to this issue is to separately adapt these VLMs to different datasets. This leads to a large number of independent VLMs, limiting their applicability to real-world applications, \eg, AI assistants, autonomous driving systems, and robotics, where versatile recognition of datasets from different domains is required.

To this end, we explore an \textbf{Open-Domain Continual Learning} (ODCL) problem for VLMs, aiming to continuously improve the recognition abilities of the pre-trained VLMs in open-domain scenarios, where the VLMs need to perform continual updating and inference on streaming of downstream datasets from diverse domains, including both \textit{seen and unseen domains} during the CL process. Due to the open nature, ODCL is also required to handle \textit{novel classes} that do not present in all historic downstream training data.

Compared to conventional continual learning (CL) \cite{chaudhry2018riemannian,bang2021rainbow,rebuffi2017icarl,douillard2020podnet,kirkpatrick2017overcoming,Dhar_2019_CVPR,yan2021dynamically,wang2022foster,zhou2022model,smith2023coda,gao2023unified}, ODCL presents two unique challenges. 1) \textbf{Strong label correlations and large domain gaps}: Unlike conventional CL
that is a closed-set task, where each task consists of \textit{known classes} (\ie, those seen in the CL process) drawn from disjoint subsets within a single dataset, ODCL can involve strongly correlated (\eg, overlapping or hierarchical) relations between open-set classes from datasets of different tasks as shown in \cref{fig:graph1}. 
Further, conventional CL predominately assumes that data comes from the same domain, while ODCL exhibits significant domain gaps across the multi-domain datasets.
2) \textbf{Two types of forgetting}: In addition to the possible forgetting of the knowledge learned in downstream tasks like conventional CL, ODCL requires the VLMs to avoid forgetting of the zero-shot recognition capabilities learned during pre-training. Without access to a diverse range of concepts similar to those encountered during pre-training, it becomes particularly challenging for the models to maintain zero-shot knowledge when learning on new datasets from different domains.

Although several recent CL methods, \eg, LAE~\cite{gao2023unified} and CODA-Prompt~\cite{smith2023coda},  leverage the pre-trained models for the CL task, they primarily adopt a single domain setting, with no overlapping classes between different tasks, rendering them ineffective for open-domain CL.
Further, they are tailored to the visual modality of the pre-training models, overlooking information from the text modality. Thus, these methods lack zero-shot capabilities on novel classes and unseen domains. 
VR-LwF~\cite{ding2022dont} and S-Prompt~\cite{wang2022sprompts} utilize both modalities of CLIP for the CL task, but they focus on different problems separately: the preservation of zero-shot recognition capabilities and a domain-incremental learning setting. ZSCL~\cite{zheng2023preventing} is the most related work, which is designed to maintain the zero-shot capabilities of CLIP while continuously learning the knowledge of downstream tasks in open environments, where novel classes and unseen domains are expected. However, it heavily relies on continuously distilling knowledge on a large external dataset to achieve this balance, which is costly in resource and computation time.

In this paper, we propose a novel \underline{co}ntinual \underline{le}arning approach based on CLIP, termed \textbf{CoLeCLIP}, which can tackle the above two open-domain CL challenges by performing a joint learning of task prompts and cross-domain class vocabulary. The cross-domain class vocabulary learning aims to preserve the text prompt embeddings of all classes across different tasks within a unified semantic space. To alleviate the forgetting of the zero-shot knowledge, our class vocabulary learning incorporates Parameter-Efficient Fine-Tuning (PEFT) modules into the frozen text encoder of CLIP to learn the text embeddings of classes in each downstream CL task separately, one independent PEFT per task, rather than fully fine-tuning the entire CLIP model as in ZSCL. Simultaneously, we also stabilize the class vocabulary by a momentum updating of newly acquired class embeddings. To tackle the challenge of large domain gaps and strong class correlations between tasks, a lightweight task prompt in the frozen image encoder of CLIP is learned for each task to capture domain-specific patterns. This is coupled with an expansion of the local class space within each task, enabling the differentiation of the correlated classes across different tasks. 

In summary, our contributions are as follows:
\begin{itemize}
\item[$\bullet$] 
We explore the problem of open-domain CL for VLMs, which emphasizes the recognition capability for known and novel classes from seen and unseen domains, while effectively preserving the pre-trained zero-shot knowledge and newly acquired knowledge from downstream CL tasks. Open-domain CL in a task-incremental learning setting was recently explored in \cite{zheng2023preventing}. We extend this problem with a class-incremental learning setting.

\item[$\bullet$] We then propose a lightweight yet effective approach CoLeCLIP for open-domain CL. It tackles the aforementioned unique challenges of this problem by jointly learning task prompts and class embeddings in the frozen image and text encoders of CLIP, respectively. This results in a parameter-efficient, rehearsal-free, and well-generalized CL approach.

\item[$\bullet$] Extensive experiments on 11 domain datasets are performed for open-domain CL under both task- and class-incremental settings, in which we show that CoLeCLIP outperforms state-of-the-art methods.
\end{itemize}

%% file: sec/2_related_work.tex
\section{Related Work}
\noindent\textbf{Continual Learning.}
CL methods can be broadly classified into three categories: rehearsal-based methods, regularization-based methods, and architecture-based methods. Rehearsal-based methods~\cite{chaudhry2018riemannian,bang2021rainbow,rebuffi2017icarl,prabhu2020gdumb,lopez2017gradient} use an external memory buffer to store a small number of exemplars from previous tasks. When learning a new task, models can revisit the stored exemplars to retain the old knowledge. Regularization-based methods~\cite{li2017learning,douillard2020podnet,kirkpatrick2017overcoming,zenke2017continual, Dhar_2019_CVPR} impose constraints on the model's parameter space or output space. As an effective strategy, knowledge distillation~\cite{hinton2015distilling} is widely employed to encourage the new model to maintain similar predicted probabilities for old classes~\cite{li2017learning,wu2019large} or feature representations akin to those of old models~\cite{douillard2020podnet,lu2022augmented}.

Architecture-based methods ~\cite{yoon2017lifelong,yan2021dynamically,wang2022foster,zhou2022model,smith2023coda,gao2023unified} expand the model structure to adapt to the new data distribution, but continuously increasing model parameters lead to additional storage costs. DER~\cite{yan2021dynamically} expands a new backbone for each task, greatly increasing memory requirements. L2P~\cite{wang2022learning}, Dualprompt~\cite{wang2022dualprompt}, CODA-Prompt~\cite{smith2023coda} and LAE~\cite{gao2023unified} fine-tune the pre-trained model through PEFT modules~\cite{houlsby2019parameterefficient,hu2021lora,li2021prefixtuning,lester2021power}, storing task knowledge in lightweight learnable parameters to reduce storage costs. Our method can further reduce the parameters required for the visual modality, as it simultaneously takes inputs from both the visual and textual modalities. Compared with DyTox~\cite{douillard2021dytox}, our method shares the pre-trained image encoder parameters across different tasks, only fine-tuning individual tokens for each task, which helps refine the original image embeddings while having substantially fewer parameters.

\noindent\textbf{Continual Learning of VLMs.}
Limited studies~\cite{wang2022sprompts,zhou2023learning,zheng2023preventing,ding2022dont} have been reported on empowering VLMs with the ability for continual learning, despite its significance to various real-world applications.
S-Prompt~\cite{wang2022sprompts} proposes a visual-language prompt learning method that simultaneously learns prompts for both the text and image encoders of CLIP for each task. However, due to the limited capability of its simple KNN operation for task identification, it fails to achieve satisfactory results in CL with different classes across tasks. Due to the open-vocabulary capability of VLMs, VR-LwF~\cite{ding2022dont} and Proof~\cite{zhou2023learning} explore how to maintain the zero-shot transfer ability of these models in CL. Proof~\cite{zhou2023learning} designs a fusion module to fully leverage knowledge from both modalities and mitigates forgetting by learning task-specific projections for each task with exemplars from previous classes. VR-LwF~\cite{ding2022dont} preserves zero-shot knowledge through distillation with replayed vocabularies. Different from these methods that focus on CL with a single-domain dataset, our work addresses an open-domain CL problem. To our knowledge, ZSCL~\cite{zheng2023preventing} is the only method designed for ODCL. It effectively alleviates the forgetting of knowledge from different domains by a knowledge distillation in the feature space and a weight ensemble updating in the parameter space. However, ZSCL requires a large reference dataset with rich semantics and fully fine-tunes the model, which significantly increases storage and training computation overhead. Compared with ZSCL, our method is very lightweight yet achieves substantially better CL performance (see Sec. \ref{subsec:main_results}).


%% file: sec/3_method.tex
\section{Our Proposed Approach: CoLeCLIP}

\subsection{Problem Statement}
In ODCL, a VLM sequentially learns $T$ tasks during training, where each task is from different datasets with large domain gaps and/or strong class correlations between them. The objective is to improve the recognition capability of the VLM for samples from seen domains while preserving the model’s zero-shot capability for unseen domains containing novel classes.  Assuming there are a set of $T$ CL tasks  $\mathcal{D} = \left \{\mathcal{D}_{1},\mathcal{D}_{2},\cdots, \mathcal{D}_{T} \right \}$, each task $\mathcal{D}_{t}$ contains $M_{t}$ samples from the class set $\mathcal{Y}_{t}$, denoted as $\mathcal{D}_{t} = {\left \{ (x_{i}, y_{i}) \right \}_{i=1}^{M_{t}} }$, $ y_{i} \in \mathcal{Y}_{t} $, where $x_{i}$ represents an image, and $y_{i}$ denotes the corresponding class label. After learning on $ \mathcal{D}_{t} $, the model needs to perform well on datasets from all seen domains $\left \{ \mathcal{D}_{1}, \mathcal{D}_{2},\cdots, \mathcal{D}_{t} \right \}$ and prevent the forgetting of zero-shot knowledge on all unseen domains $\left \{ \mathcal{D}_{t+1}, \mathcal{D}_{t+2},\cdots, \mathcal{D}_{T} \right \}$. 

To comprehensively evaluate the recognition capability of ODCL models, there are two inference settings: task-incremental and class-incremental learning, denoted as ODCL-TIL and ODCL-CIL respectively. ODCL-TIL assumes that the task identity (ID) $t_\text{id}$ can be obtained during the inference stage, while ODCL-CIL assumes that the task ID $t_\text{id}$ is not provided during inference. That is, in ODCL-TIL we are required to classify images into a particular class set $\mathcal{Y}_{t_\text{id}}$ of their domain, whereas in ODCL-CIL the predicted class space is set to be the union of the class sets $\mathcal{C}=\cup_{i=1}^{t_\text{id}}\mathcal{Y}_{i}$ from all seen tasks. 
Thus, ODCL-TIL is a relatively easier problem than ODCL-CIL. 
The ODCL-TIL setting was recently explored in ~\cite{zheng2023preventing}. We introduce the ODCL-CIL setting to enrich the evaluation of open-domain CL techniques.

\begin{figure*}[htb]
  \centering
   \includegraphics[width=0.7\linewidth]{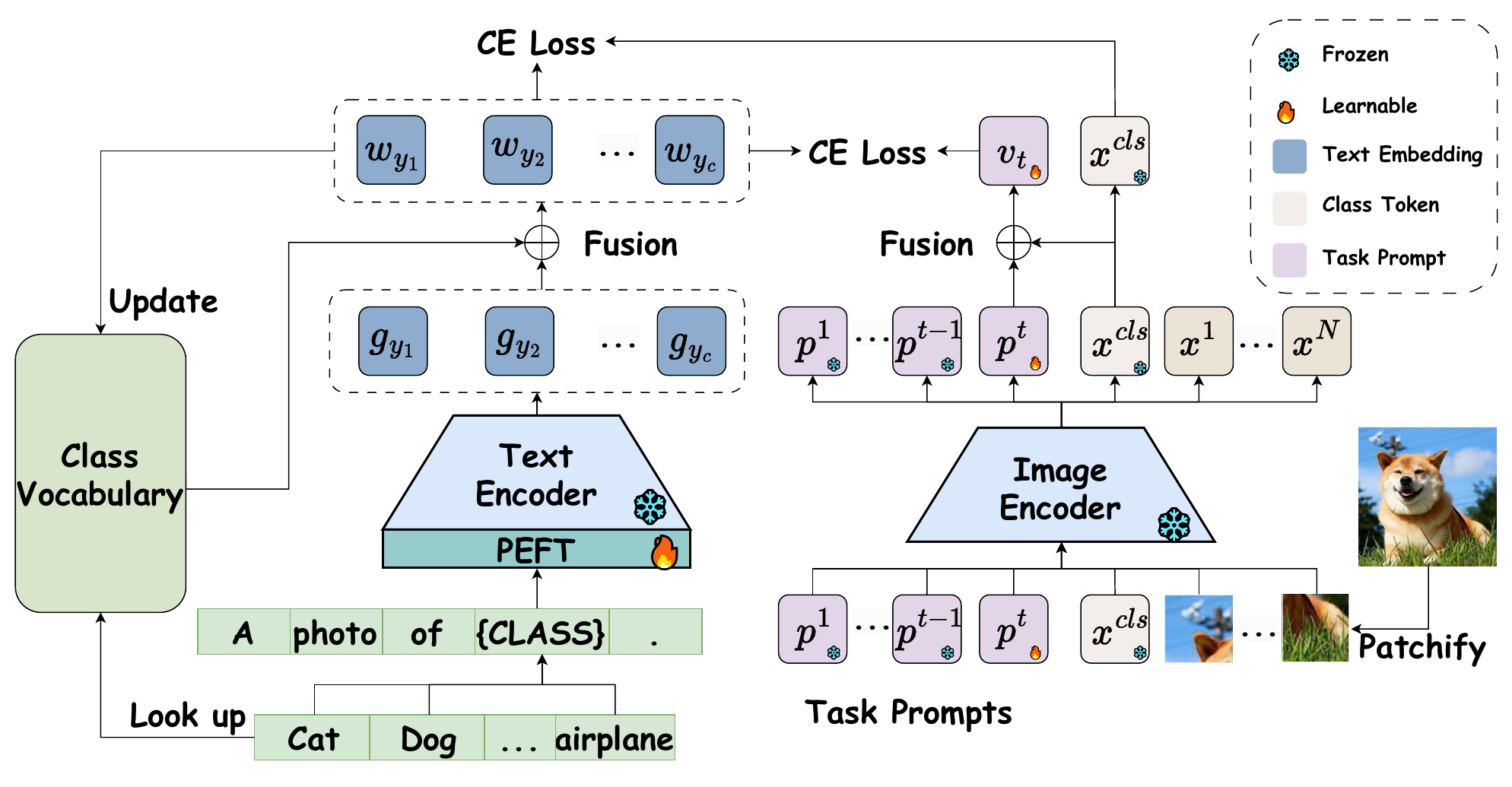}
   \caption{Illustration of the training pipeline of CoLeCLIP at a time step $t$ of the CL task. For vocabulary learning, CoLeCLIP attaches an additional PEFT module into the text encoder of CLIP to refine the text embeddings of the classes per task and builds a class vocabulary with momentum updates. For task prompt learning, CoLeCLIP leverages a task prompt for each task to extract domain-specific patterns. In the training phase, the text encoder and image encoder of CLIP are kept frozen, while only the PEFT module and a task prompt are fine-tuned. For the purpose of illustration, we omit the subscripts representing the layer number and use $g_{y}$ instead of $g _{t} (y)$. }
   \vspace{-4mm}
  \label{fig:graph2}
\end{figure*}

\subsection{Overview of CoLeCLIP}
Our proposed CoLeCLIP is a continual learning approach based on the CLIP model. 
CLIP has a dual-encoder architecture, consisting of an image encoder $f_{v}$ and a text encoder$f_{t}$. For the image encoder, an input image $x$ is first transformed into a sequence of patch embeddings with an additional class token added, forming the input of the transformer encoder $ z_{0} = [x^{\mathrm{cls}}; x^{1}; x^{2};\cdots,x^{N}] \in \mathbb{R}^{(N+1)\times D} $, where the subscript denotes the layer number. Then, the input $ z_{0} $ is passed through the transformer encoder to obtain its embeddings, with the embedding of the class token serving as the image embedding $ f_{v}(x) $. For the text encoder, we first fill a class label $ y $ into the prompt template such as ``A photo of a \{class\}” to obtain a sentence $ S $ and then extract the text embedding $ f_{t}(S) $ through the text encoder $ f_{t} $. The similarity between the embeddings from the two encoders is then used for zero-shot prediction. To simplify the notation, the text embedding  $ f_{t}(S) $ for label $ y $ is denoted as $f_{t}(y)$. The similarity $s(x, y)$ between the image $x$ and label $y$ is defined as the cosine similarity between their embeddings. The predicted result is the class with the highest similarity.  

\Cref{fig:graph2} shows the training pipeline of CoLeCLIP. In downstream task learning, CoLeCLIP introduces three modules: task prompts $\{p^i\}_{i=1}^t$, a class vocabulary $\mathcal{V}$, and a PEFT module $g$. Before training on downstream tasks, the vocabulary $\mathcal{V}$ is initialized as an empty one. CoLeCLIP jointly learns task prompts and a PEFT-based class vocabulary to tackle the open-domain CL challenges. For vocabulary learning, we fully leverage the essential role of the text encoder in learning concepts for each task. An external learnable PEFT module per task is attached to the frozen text encoder of CLIP to learn text embeddings of the classes. To maintain knowledge from both downstream tasks and pre-training, a class vocabulary progressively integrates these refined class embeddings from downstream tasks. For task prompt learning, a set of task prompts is prepended into the original embedding sequence of the images to capture domain-specific patterns for each task. At a specific time step, only the task prompt associated with the current task is learnable, while the task prompts for previous tasks and the image encoder are kept frozen.


\subsection{Replay-Free Cross-Domain Vocabulary Learning for Mitigating Knowledge Forgetting}
In CoLeCLIP, we aim to maintain a unified semantic space to effectively alleviate the forgetting of zero-shot knowledge and newly acquired knowledge from downstream tasks. To this end, we propose replay-free cross-domain vocabulary learning, which includes two key components: flexible PEFT-based class embedding learning and momentum updates of class vocabulary.

Particularly, to mitigate the knowledge forgetting caused by interference between tasks, we fine-tune an additional learnable PEFT module attached to the frozen text encoder of CLIP for each individual task, instead of fine-tuning the entire encoder continuously. This module is responsible for learning to refine the original text embeddings of classes from the vocabulary of the current task. The text encoder with a PEFT module is denoted as ${g}_{t}$. Formally, for an input class $y$, the model retrieves the corresponding text embedding $ \mathcal{V}(y)$  from the class vocabulary $\mathcal{V}$ and combines it with a learnable embedding $ {g}_{t} (y) $ from  $ {g}_{t} $ to obtain the refined embedding $w_{y}$ via:
\begin{equation}
    w_{y} = ({g}_{t} (y) + \mathcal{V} (y)) / 2.
\end{equation}

In contrast to ZSCL that requires a large reference dataset for replay-based CL, we build a class vocabulary with momentum update to maintain knowledge of all semantic concepts during downstream task learning. The keys of the vocabulary are defined by class names, and the updated knowledge is stored in the values as embedding. When a new task arises, we first extract text embedding from CLIP for classes that do not exist in the vocabulary and then add these newly obtained class name-text embedding pairs to the vocabulary in the key-value form. For each iteration $k$, we update the vocabulary as follows: 
\begin{equation}
 \mathcal{V}_{k} (y) =  w_{y} \ast  \alpha +  \mathcal{V}_{k-1} (y) \ast  \left ( 1-\alpha \right ),
\end{equation}
where $ \alpha\in [0,1] $ is the momentum coefficient. Once the training is complete, we discard the PEFT module and only store the class vocabulary for inference.

\subsection{Task Prompt Learning for Tackling Domain Differences and Class Correlations}
The large domain gaps and strong class correlations between different tasks present a unique challenge in open-domain CL.
In CoLeCLIP, we propose task prompt learning to address this challenge. We learn a task prompt for each task to capture domain-specific patterns for tackling the domain gap issue, while an expansion of the local class space within each task is also performed to enable the differentiation of the correlated classes across different tasks.

\noindent\textbf{Domain-Specific Masked Task Prompt Learning.}
We design a lightweight task prompt for each task with an attention mask to extract domain-specific visual embeddings. For the $t$-th task $ \mathcal{D}_{t} $, we prepend a set of task prompts ${ \left \{ p^{i} \right \} }_{i=1}^{t}$ into an original embedding sequence $ z_{0} $:
\begin{equation}
z_{0}^{t} = [p^{1};\cdots,p^{t};x^{\mathrm{cls}}; x^{1};\cdots x^{N}] \in \mathbb{R}^{(N+t+1)\times D} ,
\end{equation}
where only the task prompt for the current task $ {p^{t}} $ is learnable, while the remaining prompts are learned from previous $t-1$ tasks and kept frozen.

Similar to vocabulary learning, task prompt learning aims to refine image embeddings without disrupting the pre-trained visual features. An attention mask is introduced that enables the task prompt of the current to selectively attend to patch embeddings and the class token.

Additionally, to prevent interference between tasks, we impose a constraint that prohibits the preceding task prompts from attending to the subsequent task prompt. Specifically, the attention mask is designed as follows:
\begin{equation}
\mathcal{M} =\begin{bmatrix}
\mathcal{L}_{t\times t} & \mathcal{T}_{t\times \left ( N+1 \right ) } \\
\mathcal{F}_{\left ( N+1 \right )\times t  } & \mathcal{T}_{\left ( N+1 \right ) \times \left ( N+1 \right ) }
\end{bmatrix},
\end{equation}
where $\mathcal{T}_{n \times m}$ represents an $n \times m$ matrix filled with True, $\mathcal{F}_{n \times m}$ represents an $n \times m$ matrix filled with False, and $\mathcal{L}_{n \times n}$ represents an $n \times n$ lower triangular matrix where the elements above the diagonal are False and all other elements are True.
The embedding sequence $ z^{t}_{0}$ for the $t$-th task is fed into the $L$-layer vision transformer encoder, and the final embedding of an input image is extracted as follows:
\begin{equation}
z_{L}^{t} = [p_{L}^{1};\cdots,p_{L}^{t};x_{L}^{\mathrm{cls}}] \in \mathbb{R}^{(t+1)\times D},
\end{equation}

For the $t$-th task, we utilize the task prompt $p_{L}^{t}$ to enhance the class token $x_{L}^{\mathrm{cls}}$ of an input image $x$, resulting in the final visual embedding $v_{t}=(p_{L}^{t} + x_{L}^{cls})/2$.

After obtaining the image embedding and text embedding, the cosine similarity for the final visual embedding ${v}_{t}$ and the original class token $x_{L}^{\mathrm{cls}}$ is calculated respectively, which is defined as $s_{\mathrm{v}}(x,y) = \frac{{{v}_{t}}^{T}w_{y}}{\left \| {{v}_{t}}\right \| \left \|w_{y}   \right \| }$ and $s_{\mathrm{cls}} (x,y ) = \frac{{x_{L}^\mathrm{cls}}^{T}w_{y}}{\left \| x_{L}^{cls}\right \| \left \|w_{y}   \right \| }$.
As for the optimization objective, we use the standard cross-entropy loss:
\begin{equation}
 \begin{split}
\mathcal{L}_{ce} = -\frac{1}{B}\sum_{i=1}^{B} \left (log\frac{exp(s_{v} (x_{i} ,y_{i})  /\tau ) }{\sum_{j = 1}^{C_{t}} exp(s_{v} (x_{i} ,y_{j})  /\tau )}\right ) \\
-\frac{1}{B}\sum_{i= 1}^{B}\left ( log\frac{exp(s_{cls} (x_{i} ,y_{i})  /\tau ) }{\sum_{j=1}^{C_{t}} exp(s_{cls} (x_{i} ,y_{j})  /\tau )} \right )   ,
  \end{split}
\end{equation}
where $ \tau $ is the temperature, $B$ is the batch size, $C_{t}$ is the number of classes in the $t$-th task, and $ y_{i} $ corresponds to the ground truth label for image $ x_{i} $. To ensure a stable vocabulary updating, we use a simple $\ell_2$ regression loss for text embeddings, which is denoted as $\mathcal{L}_{reg} = \ell_{2} \left (w_{y}-\mathcal{V}(y)\right )$. The total loss is as follows:
\begin{equation}
\begin{aligned}
    \mathcal{L} = \mathcal{L}_{ce} + \mathcal{L}_{reg}.
\end{aligned}
\end{equation}

\noindent\textbf{Alleviating Label Correlations with Energy Score-based Negative Class Labels.}
VLMs need to further enhance their ability to differentiate the current classes of the current task in the global class semantic scope, especially when there are strong correlations between the classes of different tasks. Recent pre-trained model-based CL methods~\cite{smith2023coda, gao2023unified} use the classes of the current task as the label space to calculate the cross-entropy loss, which is called local cross-entropy. However, these methods optimize the models only based on local classes, which lacks consideration of the classes in the global class semantic space and can lead to an inability to distinguish the current categories from previous ones. Thus, we propose to expand the local class space by adding negative class labels from past tasks to enhance the model's ability to differentiate the current classes.

The selection of negative labels should be based on the semantics of images. Since the difference in energy scores~\cite{DBLP:conf/nips/LiuWOL20} of a sample between different tasks can well capture the semantic similarity between the classes across tasks, we introduce an energy score-based method for the negative class selection. Particularly, the energy score of an input image $x$ for $t$-th task $\mathcal{D}_{t}$ is calculated as follows:
\begin{equation}
E(x;t)=\tau \cdot log\sum_{i=1}^{C_{t}}e^{s_{v}(x ,y_{i} )/\tau},
\end{equation}
where $s_{v} (x ,y_{i}) $ denotes the logit for the $i$-th class of the $t$-th task. If $\mathcal{D}_{t}$ is the current task, $C_{t}$ denotes the total number of classes in $\mathcal{D}_{t}$. If $\mathcal{D}_{t}$ is a previous task, $C_{t}$ represents the number of classes in $\mathcal{D}_{t}$ that do not overlap with the classes of the current task. We evenly divide the training process into two stages based on the number of training iterations. In the first stage, we use local cross-entropy as the optimization objective to increase the energy score of the training samples of the current classes. In the second stage, for the wrong predicted samples, we add selected classes from the preceding tasks as negative classes to their local label space. Specifically, we utilize a percentage threshold $\gamma$ to select the $\gamma$-th percentile of the energy score differences between the current task and a previous task within a batch. For samples with a difference above this threshold, we expand their local class space by adding those class labels from this task as the negative classes.

\begin{table*}[htb]
    \centering
    \resizebox{0.7\linewidth}{!}{
    \begin{tabular}{c c c c c c c c c c c c c }
    \hline
        Method & Aircraft & Caltech101 & CIFAR100 & DTD & EuroSAT & Flowers & Food & MNIST & OxfordPet & Cars & SUN397 & Mean \\ \hline
        $\bm{\mathit{Last}\uparrow}$ & ~ & ~ & ~ & ~ & ~ & ~ & ~ & ~ & ~ & ~ & ~ & ~ \\ \hline
        CODA-Prompt & 12.81  & 65.38  & 63.49  & 53.83  & 56.91  & 56.64  & 83.31  & 83.99  & 76.83  & 53.65  & 75.13  & 62.00  \\ 
        LAE & 13.65  & \textbf{83.06}  & 66.88  & 59.20  & 36.61  & 48.95  & 84.08  & 94.22  & 75.44  & 42.69  & 78.69  & 62.14  \\ 
        CLIP  & 24.39  & 63.65  & 40.99  & 39.26  & 52.98  & 70.04  & 88.40  & 39.56  & 88.85  & 64.52  & 63.28  & 57.81  \\
        ZSCL & 42.96  & 62.50  & 66.05  & 55.05  & \textbf{89.17}  & 89.98  & \textbf{91.81}  & \textbf{96.32 } & 93.57  & \textbf{85.16}  & 78.58  & 77.38  \\ 
        CoLeCLIP & \textbf{48.30}  & 72.00  & \textbf{67.96}  & \textbf{72.71}  & 80.13  & \textbf{96.19}  & 90.58  & 95.25  & \textbf{94.66}  & 82.79  & \textbf{78.90}  & \textbf{79.95}  \\ \hline
        $\bm{\mathit{Forgetting}\uparrow}$  & ~ & ~ & ~ & ~ & ~ & ~ & ~ & ~ & ~ & ~ & ~ & ~ \\ \hline
        CODA-Prompt & 20.18  & 73.61  & 73.43  & 62.75  & 84.66  & 69.79  & 87.42  & 83.00  & 80.96  & 65.57  & 75.13  & 70.59  \\
        LAE & 17.07  & \textbf{85.58}  & \textbf{75.64}  & 67.70  & 64.90  & 70.45  & 87.63  & \textbf{96.64}  & 86.54  & 62.49  & 78.69  & 72.12  \\ 
        CLIP  & 24.42  & 66.26  & 48.92  & 41.03  & 53.09  & 70.04  & 88.43  & 39.56  & 88.93  & 64.53  & 63.28  & 58.95  \\ 
        ZSCL & 46.46  & 65.31  & 75.23  & 61.44  & \textbf{93.96}  & 91.19  & \textbf{91.89}  & 93.94  & 94.25  & \textbf{85.53}  & 78.58  & 79.80  \\ 
        CoLeCLIP & \textbf{48.35}  & 75.93  & 74.33  & \textbf{74.97}  & 86.86  & \textbf{96.19}  & 90.73  & 95.25  & \textbf{94.79}  & 82.81  & \textbf{78.90}  & \textbf{81.74} \\ \hline
        $\bm{\mathit{Avg}\uparrow}$ & ~ & ~ & ~ & ~ & ~ & ~ & ~ & ~ & ~ & ~ & ~ & ~ \\ \hline
        CLIP  & 24.42  & 68.21  & 52.28  & 42.14  & 53.66  & 70.27  & 88.58  & 52.20  & 89.07  & 64.55  & 65.25  & 60.97  \\ 
        ZSCL & 46.46  & 67.13  & \textbf{73.66}  & 56.93  & \textbf{78.39}  & 80.75  & \textbf{89.70}  & \textbf{72.99}  & 88.72  & 64.62  & \textbf{67.85}  & 71.56  \\ 
        CoLeCLIP & \textbf{48.35}  & \textbf{77.01}  & 72.93  & \textbf{66.83}  & 74.77  & \textbf{84.26}  & 89.62  & 72.46  & \textbf{90.67}  & \textbf{67.87}  & 66.80  & \textbf{73.78}  \\ \hline
    \end{tabular}
    }
     \caption{$\mathit{Avg}$, $\mathit{Last}$, and $\mathit{Forgetting}$ results (\%) 
     in \textbf{ODCL-CIL}.
     CODA-Prompt and LAE are not applicable to the $\mathit{Avg}$ metric.}
     \label{table:CIL}
\end{table*}

\subsection{Inference}
During inference, we extract $t$ task prompts and the original image embedding for an input image after training on $t$ tasks. For a class label, we first query the class vocabulary. If the class name is found in the vocabulary, we use the corresponding value from the vocabulary as its text embedding. Otherwise, we extract the original text embedding from the text encoder of CLIP. For a class that is present in the vocabulary, we calculate the cosine similarity between its text embedding and the corresponding task prompt to obtain the output logit for classification. For classes that overlap across different tasks, we select the maximum logit values as the final output logits. For classes that are not present in the vocabulary, we calculate the cosine similarity between its text embedding and the original image embedding to obtain the output logits.


%% file: sec/4_experiment.tex
\section{Experiments}
\begin{table*}[!ht]
    \centering
    \resizebox{0.7\linewidth}{!}{
    \begin{tabular}{c c c c c c c c c c c c c }
    \hline
        Method & Aircraft & Caltech101 & CIFAR100 & DTD & EuroSAT & Flowers & Food & MNIST & OxfordPet & Cars & SUN397 & Mean \\ \hline
        $\bm{\mathit{Last}}\uparrow$ & ~ & ~ & ~ & ~ & ~ & ~ & ~ & ~ & ~ & ~ & ~ & ~ \\ \hline
        CODA-Prompt & 43.59  & 92.97  & 74.47  & 75.21  & 92.57  & 93.07  & 90.25  & 98.61  & 92.75  & 83.80  & 78.75  & 83.28  \\ 
        LAE & \textbf{49.14}  & 93.84  & 80.53  & 75.43  & 90.63  & 87.07  & 88.94  & 98.64  & 92.94  & 81.36  & \textbf{80.08}  & 83.51  \\ 
        CLIP  & 24.42  & 87.79  & 67.36  & 45.11  & 54.65  & 70.53  & 88.70  & 59.43  & 89.13  & 64.56  & 65.45  & 65.19  \\ 
        ZSCL & 42.99  & 92.17  & \textbf{81.11}  & 70.00  & \textbf{94.72}  & 90.18  & \textbf{91.84}  & \textbf{98.75}  & 93.87  & \textbf{85.18}  & 80.06  & 83.71  \\ 
        CoLeCLIP & 48.51  & \textbf{94.93}  & 79.18  & \textbf{78.03}  & 88.35  & \textbf{96.31}  & 91.18  & 97.64  & \textbf{94.90}  & 82.84  & 79.83  & \textbf{84.70}  \\ \hline
        $\bm{\mathit{Forgetting}\uparrow}$  & ~ & ~ & ~ & ~ & ~ & ~ & ~ & ~ & ~ & ~ & ~ & ~ \\ \hline
        CODA-Prompt & 47.49  & 92.84  & 78.23  & 76.40  & 94.51  & 95.23  & 90.20  & 98.55  & 93.12  & 83.99  & 78.75  & 84.48  \\ 
        LAE & \textbf{53.69}  & 94.06  & 80.50  & 76.72  & 93.54  & 90.34  & 89.46  & \textbf{98.97}  & 93.42  & 84.38  & \textbf{80.08}  & 85.02  \\
        CLIP  & 24.42  & 87.79  & 67.36  & 45.11  & 54.65  & 70.53  & 88.70  & 59.43  & 89.13  & 64.56  & 65.45  & 65.19  \\ 
        ZSCL & 46.47  & 92.53  & \textbf{82.48}  & 71.34  & \textbf{96.07}  & 91.37  & \textbf{91.90}  & 98.86  & 94.45  & \textbf{85.54}  & 80.06  & 84.64  \\ 
        CoLeCLIP & 48.51  & \textbf{95.06}  & 79.44  & \textbf{78.03}  & 92.75  & \textbf{96.31}  & 91.18  & 97.64  & \textbf{94.90}  & 82.84  & 79.83  & \textbf{85.14} \\ \hline
        $\bm{\mathit{Avg}}\uparrow$ & ~ & ~ & ~ & ~ & ~ & ~ & ~ & ~ & ~ & ~ & ~ & ~ \\ \hline
        CLIP & 24.42  & 87.79  & 67.36  & 45.11  & 54.65  & 70.53  & 88.70  & 59.43  & 89.13  & 64.56  & 65.45  & 65.19  \\ 
        ZSCL & 46.47  & 91.89  & \textbf{79.59}  & 64.13  & \textbf{79.73}  & 80.85  & 89.70  & \textbf{74.78}  & 88.78  & 64.62  & \textbf{67.99}  & 75.32  \\ 
        CoLeCLIP & \textbf{48.51}  & \textbf{94.40}  & 77.10  & \textbf{69.05}  & 78.52  & \textbf{84.32}  & \textbf{89.82}  & 73.32  & \textbf{90.70}  & \textbf{67.88}  & 66.89  & \textbf{76.41}  \\ \hline
    \end{tabular}
    }
    \caption{$\mathit{Avg}$, $\mathit{Last}$, and $\mathit{Forgetting}$ results (\%) 
    in \textbf{ODCL-TIL}.
    CODA-Prompt and LAE are not applicable to the $\mathit{Avg}$ metric.}
    \label{table:TIL}
\end{table*}

\subsection{Experimental Setup}

\noindent \textbf{Datasets.}
A benchmark consisting of 11 datasets from diverse domains, including Aircraft~\cite{maji2013fine}, Caltech101~\cite{fei2004learning}, CIFAR100~\cite{krizhevsky2009learning}, DTD~\cite{cimpoi2014describing}, EuroSAT~\cite{helber2019eurosat}, Flowers~\cite{nilsback2008automated}, Food~\cite{bossard2014food}, MNIST~\cite{deng2012mnist}, OxfordPet~\cite{parkhi2012cats}, StanfordCars~\cite{krause20133d}, and SUN397~\cite{xiao2010sun}, is used. 
Each of these datasets corresponds to a new dataset given in one of the streaming tasks in CL. There are two such streaming orders used in ~\cite{zheng2023preventing}: Order-I and Order-II. 
In Order-I, the streaming order is alphabetical w.r.t. the dataset name, while in Order-II the datasets arrive in a specific random order. 
We follow the same streaming orders as in ~\cite{zheng2023preventing}. The results we reported below are all based on Order I. The results for Order II are provided in our Appendix. After training on all 11 tasks, the class space comprises a total of 1,176 distinct classes.

\noindent \textbf{Implementation Details.} Following previous work ~\cite{zheng2023preventing}, we adopt CLIP-ViT-B/16 as an instance of VLMs that requires continual learning. We employ the Adam optimizer and a constant learning rate of 0.001 for all tasks. The batch size is set to 128 by default. To align with the number of iterations used in ZSCL and maintain the completeness of the training epochs, we round up the number of epochs separately for each dataset. The specific number of epochs assigned to each dataset can be found in our Appendix.

In our vocabulary learning, we use the LoRA-based method ~\cite{hu2021lora} for the added PEFT module and insert it into each layer in the transformer attention blocks of the CLIP text encoder. Following LAE ~\cite{gao2023unified}, the rank of LoRA is set to 5. 
Our task prompt learning is performed in the image encoder, with the length of the prompt for each task set to one and the prompt is only added to the input sequence of the transformer encoder.  The momentum coefficient $\alpha $ is set to 0.1 and the percentage threshold $\gamma$ is set to 0.7. Following CLIP ~\cite{radford2021learning}, the temperature $\tau$ is set to 0.01.

\noindent \textbf{Evaluation Metrics.}
To effectively evaluate the performance of open-domain CL,
following \cite{zheng2023preventing}, three performance metrics are used, namely $Avg$, $Last$, and $\mathit{Transfer}$, to evaluate the model's ability to preserve knowledge from the pre-training and downstream tasks. 
For brevity, we use $A_{t}$ and $T_{t}$ to represent the $ Avg $ and $\mathit{Transfer}$ metrics for the $t$-th dataset, respectively. Particularly, $Avg$ measures the average accuracy of the open-domain CL model over all datasets throughout all time steps, \ie, for each time step, we evaluate the accuracy of a model on all previously seen and emerging unseen domain datasets.  Formally, let $A_{t}^{i}$ be the accuracy of the $t$-th domain dataset after the model being trained on the task at time step $i$, then we have $A_{t} = \frac{1}{T} \sum_{i=1}^{T}A_{t}^{i}$ and $Avg = \frac{1}{T}\sum_{t=1}^{T}A_t$, where $T$ is the total number of tasks we have.
$Last$ is defined as the average accuracy across all domain datasets at the last time step.
Thus, large $Avg$ and $Last$ indicates better adaptation to the seen domains while preventing the forgetting of knowledge learned from these seen domains, as well as better preservation of CLIP's zero-shot recognition ability to emerging unseen domains. 
On the other hand, to exclusively evaluate the zero-shot performance on unseen domains, $\mathit{Transfer}$ is defined as: $\mathit{Transfer}=\frac{1}{T}\sum_{t=1}^{T}T_{t}$, where $T_{t} = \frac{1}{t-1} \sum_{i=1}^{t-1}A_{t}^{i}$, focusing on the accuracy on the unseen domain datasets only.

Additionally, we introduce a measure called $\mathit{Forgetting}$, focusing exclusively on the prevention of catastrophic forgetting of knowledge learned from the seen domains. For brevity, we use $F_{t}$ to represent the $\mathit{Forgetting}$ metrics for the $t$-th dataset.
Formally, it is defined as the average accuracy over all datasets spanning from the learned time step to the last time step: $\mathit{Forgetting}=\frac{1}{T}\sum_{t=1}^{T}F_{t}$, where $F_{t} = \frac{1}{T-t+1} \sum_{i=t}^{T}A_{t}^{i}$.

\noindent \textbf{Comparison Methods.}
CoLeCLIP is compared with ZSCL \cite{zheng2023preventing}, currently the only method designed specifically for open-domain CL. Two recent CL methods based on pre-trained models, CODA-Prompt~\cite{smith2023coda} and LAE \cite{gao2023unified}, are included as representative CL methods from a closed-domain setting for comparison. Considering that traditional CL methods do not consider non-overlapping classes between tasks, we make a simple modification to these methods. Each category is assigned a linear classifier. In the case of overlapping classes, subsequent tasks optimize the existing linear classifiers inherited from the preceding tasks, rather than introducing new ones. As CODA-Prompt and LAE are designed for the visual modality only, their model lacks zero-shot capability, and thus, they do not have \textit{Avg} and \textit{Transfer} results. The comparison to these methods is focused on only the $Last$ and $\mathit{Forgetting}$ metrics. 

\subsection{Main Results}\label{subsec:main_results}

\noindent\textbf{Performance on Seen Domains.} 
The performance of our method CoLeCLIP and its competing methods on seen domain datasets is measured by \textit{Last} and \textit{Forgetting} in Tabs. \ref{table:CIL} and \ref{table:TIL}, showing the results 
in ODCL-CIL and ODCL-TIL,
respectively. The absence of the task ID during inference makes ODCL-CIL more challenging than ODCL-TIL, and as a result, all methods in ODCL-CIL exhibit a certain degree of performance decline in both measures, compared to the results in ODCL-TIL. Nevertheless, CoLeCLIP is the best performer in the mean results under two settings, especially for ODCL-CIL where CoLeCLIP obtains about 2\% improvement over ZSCL, more than 20\% over CLIP, and about 10\%-18\% over CODA-Prompt and LAE. Particularly, as shown in Tab. \ref{table:CIL}, the final accuracy of ZSCL, \textit{Last}, on the Caltech101 dataset is lower than the zero-shot accuracy of CLIP, suggesting that fine-tuning VLMs in domain-specific space as in ZSCL may potentially disrupt the joint multi-modal embedding space.
For earlier tasks, CoLeCLIP effectively mitigates significant forgetting without a sharp decline, contrasting to that of CODA-Prompt and LAE, particularly on the first dataset. For tasks with notable domain gaps such as DTD, EuroSAT, and MNIST, due to the task prompt learning, CoLeCLIP is able to achieve a better balance between previous knowledge and newly acquired knowledge. 

\begin{table*}[!ht]
    \centering
    \resizebox{0.8\linewidth}{!}{
    \begin{tabular}{c c c c c c c c c c c c c }
    \hline
        Method & Caltech101 & CIFAR100 & DTD & EuroSAT & Flowers & Food & MNIST & OxfordPet & Cars & SUN397 & Mean \\ \hline
        CLIP  & \textbf{87.79}  & \textbf{67.36}  & \textbf{45.11}  & \textbf{54.65}  & \textbf{70.53}  & \textbf{88.70}  & 59.43  & \textbf{89.13}  & \textbf{64.56}  & 65.45  & \textbf{69.27}  \\ 
        ZSCL & 85.43  & 66.58  & 44.93  & 51.13  & 68.22  & 87.87  & \textbf{61.02}  & 86.65  & 59.97  & \textbf{66.78}  & 67.86  \\ 
        CoLeCLIP & \textbf{87.79}  & 66.60  & \textbf{45.11}  & 53.61  & 69.94  & \textbf{88.70}  & 59.43  & \textbf{89.13}  & \textbf{64.56}  & 65.59  & 69.04 \\ \hline
    \end{tabular}
    }
    \caption{$\mathit{Transfer}$ results (\%) of CoLeCLIP and its two competing methods that can work on unseen domains. 
    Note that the results are applicable to ODCL-TIL only as we need to know the task IDs to justify whether they are seen or unseen domains.
    }
    \label{table:t}
\end{table*}

As shown in \cref{table:TIL}, in ODCL-TIL,
by distilling in the feature space with a large reference dataset and weight ensemble updates, ZSCL can maintain the knowledge from different downstream domain datasets, with good performance in both metrics across all datasets. Without the assistance of additional datasets, CODA-Prompt and LAE can still achieve comparably good performance to ZSCL in \textit{Last} by storing task knowledge in prompts or additional PEFT modules. Our CoLeCLIP also falls under the category of replay-free approaches as CODA-Prompt and LAE, but it is much more effective: it obtains the best \textit{Last} and \textit{Forgetting} averaged over all the datasets, due to its excellent performance on selected datasets like DTD and Flowers, significantly outperforming ZSCL by 5\%-10\% in both measures, while maintaining similarly good performance as ZSCL on the other datasets.

\noindent\textbf{Performance on Both Seen and Unseen Domains.} 
The \textit{Avg} metric evaluates the ability of CL methods to balance the accuracy on seen domains and the generalization to unseen domains in open-domain CL. As shown in Tabs. \ref{table:CIL} and \ref{table:TIL}, CLIP excels in the zero-shot generalization to unseen domains but performs poorly on the seen domains due to the lack of adaptation to the downstream tasks. Our method CoLeCLIP significantly improves CLIP in seen domain performance, while effectively preserving its zero-shot capabilities, achieving a better balance between learning the seen-domain tasks and preserving knowledge from both seen and unseen domains. As a result, it obtains the best \textit{Avg} performance 
in both ODCL-CIL and ODCL-TIL settings, surpassing ZSCL by more than 2\% in ODCL-CIL and 1\% in ODCL-TIL.

\noindent\textbf{Generalization to Unseen Domains.} 
Tab. \ref{table:t} shows the \textit{Transfer} results of the three methods that have zero-shot recognition capabilities. 
ZSCL performs fairly well in maintaining the zero-shot accuracy of CLIP, with only a 1.41\% decline. CoLeCLIP maintains a unified class semantic space through class vocabulary learning, which helps enhance the zero-shot capabilities for samples from unseen domains, further reducing the zero-shot accuracy decline from 1.41\% in ZSCL to 0.23\%. This demonstrates the substantial improvement of CoLeCLIP over ZSCL in avoiding the forgetting of pre-trained zero-shot knowledge.

\subsection{Analysis of CoLeCLIP}

\noindent\textbf{Ablation Study.} 
The ablation study results of CoLeCLIP and its six variants 
in ODCL-CIL
are reported in \cref{table:abl}, where the first row is a baseline representing a simplified use of our vocabulary learning that learns the text embedding for classes of downstream tasks based on LoRA separately without class embedding updating. The baseline achieves better \textit{Transfer} than ZSCL, but it exhibits higher forgetting of knowledge learned from different tasks, highlighted by its \textit{Last} being almost 10\% lower than ZSCL. With the momentum updating of class embeddings through our class vocabulary in the second row, it effectively avoids disrupting the unified semantic space, better preserving domain-specific knowledge from the seen domain as well as zero-shot knowledge. This helps improve \textit{Last} by 8.48\%, as well as a slight increase in \textit{Transfer}.

Using the task prompt learning alone can effectively capture domain-specific patterns,
alleviating the catastrophic forgetting problem intensified by the large domain gaps, as manifested by an increased performance in \textit{Last}. As shown in the penultimate row of \cref{table:abl}, our task prompt learning further enhances the discriminative ability of the current categories across all categories from the seen domain by additionally selecting negative labels to expand the local label space, resulting in an improvement in \textit{Last}. This is done to increase the penalty for misclassified samples, even when there are strong category correlations among these tasks. Moreover, we observe that the independent use of negative class label selection or its combination with the class vocabulary can also enhance the discriminative capability for categories from all seen domains, as illustrated by the improved \textit{Last} performance. The combination of these three components yields the best performance across \textit{Avg}, \textit{Last}, and \textit{Forgetting}, with only a minor drop in \textit{Transfer}.

\begin{table}[!t]
    \centering
    \resizebox{0.8\linewidth}{!}{
    \begin{tabular}{c c c|c c c c }
    \hline
        \thead{Vocabulary \\ Learning} & \thead{Masked Task \\ Prompts} & \thead{Negative Class \\ Label Selection}  
        & Transfer$\uparrow$ & Avg$\uparrow$ & Last$\uparrow$ & Forgetting$\uparrow$ \\ \hline
        - & - & - & 68.95  & 67.48  & 67.73  & 68.46  \\ \hline
        \checkmark & ~ & ~ & 69.24  & 71.88  & 76.21  & 77.98  \\ 
        ~ & \checkmark & ~ & 69.03  & 71.52  & 77.07  & 78.12  \\ 
        ~ & ~ & \checkmark & 68.84  & 67.92  & 68.56  & 69.27  \\ 
        \checkmark & \checkmark & ~ & 69.04  & 73.62  & 79.24  & 81.52  \\ 
        \checkmark & ~ & \checkmark & \textbf{69.25}  & 72.24  & 76.98  & 78.69  \\ 
        ~ & \checkmark & \checkmark & 69.04  & 71.99  & 78.00  & 78.86  \\ 
        \checkmark & \checkmark & \checkmark & 69.04  & \textbf{73.78}  & \textbf{79.95}  & \textbf{81.74} \\ \hline
    \end{tabular}
    }
    \caption{Ablation study results of the variants of CoLeCLIP.}
    \label{table:abl}
\end{table}

\noindent \textbf{The Use of Alternative PEFT Modules.} In addition to the default LoRA module used in the PEFT module for the text encoder, we also explored two other commonly used PEFT modules -- Adapter ~\cite{houlsby2019parameterefficient} and Prefix ~\cite{li2021prefixtuning} -- to learn the class embeddings in the vocabulary. As shown in \cref{table:PEFT}, LoRA works best in all three metrics 
in both ODCL-TIL and ODCL-CIL, followed by Adapter and Prefix.

\begin{table}[!t]
    \centering
    \resizebox{0.8\linewidth}{!}{
    \begin{tabular}{c|c|ccc|ccc}
    \hline
        \multirow{2}{*}{PEFT} & \multirow{2}{*}{Transfer} & \multicolumn{3}{c|}{ODCL-TIL} & \multicolumn{3}{c}{ODCL-CIL} \\ \cline{3-8}
        ~ & ~ &Avg$\uparrow$ & Forgetting$\uparrow$ & Last$\uparrow$  & Avg$\uparrow$ & Forgetting$\uparrow$ & Last$\uparrow$  \\ \hline
        Prefix & 68.28  & 75.43  & 83.88  & 83.61  & 72.34  & 79.60  & 77.19  \\ 
        Adapter & 68.61  & 76.19  & 84.92  & 84.66  & 73.56  & 81.34  & 78.89  \\ 
        LoRA & \textbf{69.04}  & \textbf{76.41}  & \textbf{85.14}  & \textbf{84.70}  & \textbf{73.78}  & \textbf{81.74}  & \textbf{79.95} \\ \hline
    \end{tabular}
    }
    \caption{Using different PEFT methods in the text encoder.}
    \label{table:PEFT}
\end{table}

\begin{table}[!t]
    \centering
     \resizebox{0.8\linewidth}{!}{
    \begin{tabular}{c|c c c c c}
    \hline
        Method & Batch Size & \thead{\# Parameters (M)} & Training (s) & \thead{Memory (G)} & Inference (s) \\ \hline
        CODA-Prompt  & 128 & 9.51 & 43.30 & \textbf{24.7} & 34.98 \\ 
        LAE  & 128 & 0.24 & \textbf{34.02} &  25.4 & \textbf{27.18} \\ 
        ZSCL & 64 & 149.62 & 256.18 & 43.3 & 27.51 \\ 
        CoLeCLIP & 128 & \textbf{0.13} & 37.38 &  26.8 & 28.86 \\ \hline
    \end{tabular}
    }
     \caption{Computation overhead and memory usage.}
      \label{table:overhead}
\end{table}

 \noindent \textbf{Time and Space Overhead.}
\cref{table:overhead} shows the computation time and GPU memory usage of CoLeCLIP on the Aircraft dataset, with the other three methods as baselines. `Training' represents the training time of a single epoch, while `Inference' represents the time required to perform inference on the full Aircraft test set. Because LAE, CODA-Prompt, and CoLeCLIP adopt PEFT modules to fine-tune CLIP, they can effectively save GPU memory and accelerate training, compared to the fully fine-tuned method ZSCL. These methods save nearly half of the GPU memory and reduce the training time by more than five times even when the batch size is doubled.  More importantly, CoLeCLIP has the smallest number of learnable parameters. During inference, CoLeCLIP and ZSCL have comparable efficiency;
CODA-Prompt introduces additional computation overhead due to its prompt generation at each layer. Overall, although CoLeCLIP has lightweight requirements, it is the most effective method (as shown in Tabs. \ref{table:CIL}, \ref{table:TIL}, and \ref{table:t}).


%% file: sec/5_conclusion.tex
\section{Conclusion}
In this work, we explore the problem of ODCL for VLMs, which necessitates VLMs to learn the recognition capabilities for the classes in the adapted domain datasets while maintaining zero-shot knowledge to unseen domains with novel classes. 
We propose a lightweight and effective method CoLeCLIP alleviating the forgetting of the pre-trained zero-shot knowledge and the new knowledge acquired from downstream tasks through a joint learning of task prompts and a cross-domain class vocabulary. 
In both ODCL-CIL and ODCL-TIL settings,
CoLeCLIP outperforms state-of-the-art methods. Overall, ODCL is a largely unexplored but practically-important problem. CoLeCLIP provides a cost-effective solution to tackle the problem.

%% file: sec/X_suppl.tex
\clearpage
\appendix
\setcounter{page}{1}
\maketitlesupplementary

\section{Additional Implementation Details}
 \noindent \textbf{Training Epochs.} The number of epochs assigned to each dataset is set as follows: Aircraft~\cite{maji2013fine} (20 epochs), Caltech101~\cite{fei2004learning} (10 epochs), CIFAR100~\cite{krizhevsky2009learning} (2 epochs), DTD~\cite{cimpoi2014describing} (35 epochs), EuroSAT~\cite{helber2019eurosat} (3 epochs), Flowers~\cite{nilsback2008automated} (63 epochs), Food~\cite{bossard2014food} (1 epoch), MNIST~\cite{deng2012mnist} (2 epochs), OxfordPet~\cite{parkhi2012cats} (18 epochs), StanfordCars~\cite{krause20133d} (8 epochs), and SUN397~\cite{xiao2010sun} (1 epoch).

 \noindent \textbf{Comparison Methods.} The comparative methods, CODA-Prompt~\cite{smith2023coda} and LAE~\cite{gao2023unified}, employ a similar training strategy as our method to adapt to the open-domain CL problem. Specifically, CODA-Prompt uses the Adam optimizer with a constant learning rate of 0.01 for all tasks, while LAE uses the Adam optimizer with a constant learning rate of 0.001 for all tasks. The batch size for both methods is set to 128. For LAE, we use the LoRA-based method~\cite{hu2021lora} for the PEFT module and insert these modules into the attention blocks of the image encoder of CLIP~\cite{radford2021learning} in each layer. Following LAE, the rank of LoRA is set to 5, the weight decay of the EMA algorithm is set to 0.9999 and the number of iterations for freezing PEFT modules is set to 60\% of the total number of iterations. For CODA-Prompt, we follow its hyperparameter choices.  The prompt length is set to 8, and the length of a set of prompt components is set to 110 due to a total of 11 tasks. Similar to LAE, we use prompts in each layer of the image encoder of CLIP. The weight for the orthogonality loss is set to 0.1. For both methods, they use the same number of training epochs for each dataset as our method. 
 
 \noindent \textbf{The Hyperparameter Settings for Alternative PEFT Modules.} Apart from the explanations for the specific hyperparameters below, the rest of the training strategy and hyperparameter selections remain unchanged. For the Prefix-based PEFT modules~\cite{li2021prefixtuning}, the length of the prefix is set to 10, and the learning rate is set to 0.01. For the Adapter-based PEFT modules~\cite{houlsby2019parameterefficient}, the down-projection dimension of the adapter is set to 5.

\section{Additional Results for Order I}
\noindent \textbf{Analysis of Hyperparameter Settings.} \cref{table:hyper} showcases the impact of hyperparameters on the performance of CoLeCLIP under both ODCL-TIL and ODCL-CIL settings. \cref{table:alpha} reveals that within an appropriate range, \eg, [0.01,0.1], different values of the momentum coefficient $\alpha$ have a relatively minor impact on the performance. However, when the momentum coefficient is excessively small, \eg, 0.001, the updates of category vocabulary become extremely slow, which can result in a decline in the performance on seen domains. \cref{table:gamma} indicates that CoLeCLIP can perform stably w.r.t. a wide range of the settings of $\gamma$.

\begin{table}[htb]
  \centering
  \begin{subtable}{0.9\linewidth}
    \centering
     \resizebox{\linewidth}{!}{
    \begin{tabular}{c|c|c c c|c c c}
    \hline
        \multirow{2}{*}{$\alpha$} & \multirow{2}{*}{$\bm{\mathit{Transfer}}\uparrow$} & \multicolumn{3}{c|}{ODCL-TIL} & \multicolumn{3}{c}{ODCL-CIL} \\ \cline{3-8}
        ~ & ~ & $\bm{\mathit{Avg}}\uparrow$ & $\bm{\mathit{Forgetting}}\uparrow$ & $\bm{\mathit{Last}}\uparrow$ &$\bm{\mathit{ Avg}}\uparrow$ & $\bm{\mathit{Forgetting}}\uparrow$ & $\bm{\mathit{Last}}\uparrow$ \\ \hline
        0.1 & 69.04  & \textbf{76.41}  & 85.14  & 84.70  & \textbf{73.78}  & \textbf{81.74}  & 79.95  \\ 
        0.05 & 69.18  & \textbf{76.41}  & 85.07  & 84.66  & 73.30  & 81.05  & 79.15  \\ 
        0.01 & 68.98  & 76.40  & \textbf{85.19}  & \textbf{84.73}  & 73.54  & 81.43  & \textbf{79.97}  \\ 
        0.005 & 69.03  & 76.07  & 84.68  & 84.49  & 73.00  & 80.82  & 78.87  \\
        0.001 & \textbf{69.24}  & 73.44  & 80.14  & 80.06  & 69.85  & 75.55  & 73.40 \\ \hline
    \end{tabular}
    }
        \caption{Momentum coefficient $\alpha$.}
    \label{table:alpha}
  \end{subtable}
  \quad
  \begin{subtable}{0.9\linewidth}
    \resizebox{\linewidth}{!}{
    \begin{tabular}{c|c|c c c|c c c}
    \hline
        \multirow{2}{*}{$\gamma$} & \multirow{2}{*}{$\bm{\mathit{Transfer}}$} & \multicolumn{3}{c|}{ODCL-TIL} & \multicolumn{3}{c}{ODCL-CIL} \\ \cline{3-8}
        ~ & ~ & $\bm{\mathit{Avg}}\uparrow$ & $\bm{\mathit{Forgetting}}\uparrow$ & $\bm{\mathit{Last}}\uparrow$ & $\bm{\mathit{Avg}}\uparrow$ & $\bm{\mathit{Forgetting}}\uparrow$ & $\bm{\mathit{Last}}\uparrow$ \\ \hline
        0.1 & 68.99  & 76.26  & 85.02  & 84.47  & 73.60  & 81.67  & 79.81  \\
        0.3 & 69.02  & 76.32  & 85.02  & 84.57  & 73.70  & 81.71  & 79.91  \\ 
        0.5 & \textbf{69.04}  & 76.34  & 85.04  & 84.61  & 73.75  & 81.70  & \textbf{79.99}  \\ 
        0.7 & \textbf{69.04}  & \textbf{76.41}  & 85.14  & 84.70  & 73.78  & 81.74  & 79.95  \\ 
        0.9 & \textbf{69.04}  & \textbf{76.41}  & \textbf{85.15}  & \textbf{84.79}  & \textbf{73.81}  & \textbf{81.77}  & 79.92 \\ \hline
    \end{tabular}
    }
    \caption{The percentage threshold $\gamma$.}
    \label{table:gamma}
  \end{subtable}
  \caption{Empirical analysis of hyperparameters $\alpha$ and $\gamma$ on the CoLeCLIP performance for Order I.}
  \label{table:hyper}
\end{table}

\section{Results for Order II}
The results of open-domain CL under Order II for the streaming CL tasks are shown in Tabs. \ref{table:CIL_order2} and \ref{table:TIL_order2} for ODCL-CIL and ODCL-TIL, respectively. Below we analyze the results in detail.

\noindent \textbf{Performance on Seen Domains.} Consistent with the findings in Order I, CoLeCLIP exhibits the best performance in terms of $\mathit{Last}$ and $\mathit{Forgetting}$ results in \cref{table:CIL_order2} under the ODCL-CIL setting. It shows an improvement of approximately 2-3\% compared to ZSCL~\cite{zheng2023preventing}, surpassing CLIP by over 20\%, and achieving enhancements of around 8\%-16\% over CODA-Prompt and LAE. CoLeCLIP is capable of retaining expert knowledge from earlier tasks, achieving the highest or second-highest $\mathit{Last}$ and $\mathit{Forgetting}$ results in the first seven tasks. Moreover, CoLeCLIP effectively mitigates the forgetting of knowledge from domains with significant domain gaps. For example, CoLeCLIP achieves relatively balanced performance in terms of $\mathit{Last}$ and $\mathit{Forgetting}$ metrics in datasets such as DTD, EuroSAT, and MNIST. As shown in \cref{table:TIL_order2}, CoLeCLIP achieves the highest $\mathit{Last}$ metric and comparable $\mathit{forgetting}$ metric with a marginal difference (approximately 0.08\%) under the ODCL-TIL setting. Similar to the conclusion in Order I, CoLeCLIP outperforms ZSCL by 4\%-9\% in both measures on the DTD and Flowers datasets, while maintaining comparable performance to ZSCL on multiple other datasets.

\begin{table*}[!ht]
    \centering
    \resizebox{0.8\linewidth}{!}{
    \begin{tabular}{c c c c c c c c c c c c c}
    \hline
        Method & Cars & Food & MNIST & OxfordPet & Flowers & SUN397 & Aircraft & Caltech101 & DTD & EuroSAT & CIFAR100 & Mean \\ \hline
        $\bm{\mathit{Last}}\uparrow$ & ~ & ~ & ~ & ~ & ~ & ~ & ~ & ~ & ~ & ~ & ~ & ~ \\ \hline
        CODA-Prompt & 53.35  & 78.54  & 83.12  & 79.97  & 45.80  & 71.14  & 33.54  & 71.49  & 59.10  & 53.41  & 75.71  & 64.11  \\
        LAE & 75.70  & 82.77  & 78.51  & 80.19  & 27.08  & 68.56  & 7.59  & \textbf{75.69}  & 59.31  & 47.96  & \textbf{83.81}  & 62.47  \\
        CLIP & 64.52  & 88.40  & 39.56  & 88.85  & 70.04  & 63.28  & 24.39  & 63.65  & 39.26  & 52.98  & 40.99  & 57.81  \\ 
        ZSCL & 77.89  & \textbf{91.11}  & 89.18  & 92.64  & 85.84  & 76.75  & 45.57  & 60.66  & 58.62  & \textbf{85.26}  & 74.93  & 76.22  \\ 
        CoLeCLIP  & \textbf{82.75}  & 90.48  & \textbf{92.04}  & \textbf{94.19}  & \textbf{95.79}  & \textbf{77.52}  & \textbf{48.36}  & 69.87  & \textbf{71.65}  & 79.96  & 71.15  & \textbf{79.43}  \\ \hline
        $\bm{\mathit{Forgetting}}\uparrow$  & ~ & ~ & ~ & ~ & ~ & ~ & ~ & ~ & ~ & ~ & ~ & ~ \\ \hline
        CODA-Prompt & 68.54  & 83.26  & 94.59  & 84.93  & 69.72  & 76.79  & 38.08  & 82.62  & 63.55  & 58.54  & 75.71  & 72.39  \\
        LAE & \textbf{84.00}  & 86.58  & 94.48  & 84.62  & 60.18  & 75.87  & 13.93  & \textbf{86.75}  & 68.44  & 71.17  & \textbf{83.81}  & 73.62  \\ 
        CLIP & 64.55  & 88.60  & 48.18  & 88.97  & 70.44  & 64.33  & 24.39  & 68.99  & 39.68  & 53.02  & 40.99  & 59.29  \\ 
        ZSCL & 81.55  & \textbf{91.56}  & 95.81  & 93.35  & 90.33  & 77.69  & \textbf{49.01}  & 71.41  & 62.91  & 79.55  & 74.93  & 78.92  \\ 
        CoLeCLIP  & 83.23  & 90.92  & \textbf{96.68}  & \textbf{94.34}  & \textbf{96.18}  & \textbf{78.66}  & 48.42  & 78.34  & \textbf{72.13 } & \textbf{84.91}  & 71.15  & \textbf{81.36}  \\ \hline
        $\bm{\mathit{Avg}}\uparrow$ & ~ & ~ & ~ & ~ & ~ & ~ & ~ & ~ & ~ & ~ & ~ & ~ \\ \hline
        CLIP & 64.55  & 88.61  & 50.23  & 89.01  & 70.47  & 64.84  & 24.41  & 80.95  & 43.63  & 54.35  & 64.96  & 63.27  \\ 
        ZSCL & 81.55  & \textbf{91.28}  & 88.89  & 91.09  & 81.71  & 71.91  & 33.75  & 82.13  & 50.32  & 59.19  & \textbf{68.43}  & 72.75  \\
        CoLeCLIP  & \textbf{83.23}  & 90.72  & \textbf{89.91} & \textbf{92.92}  & \textbf{86.85}  & \textbf{72.66}  & \textbf{35.33}  & \textbf{83.85}  & \textbf{52.48}  & \textbf{59.63}  & 61.78  & \textbf{73.58} \\ \hline
    \end{tabular}
    }
     \caption{$\mathit{Avg}$, $\mathit{Last}$, and $\mathit{Forgetting}$ results (\%) in \textbf{ODCL-CIL} for Order II. CODA-Prompt and LAE are not applicable to $\mathit{Avg}$.}
    \label{table:CIL_order2}
\end{table*}

\begin{table*}[!ht]
    \centering
     \resizebox{0.8\linewidth}{!}{
    \begin{tabular}{c c c c c c c c c c c c c}
    \hline
        Method & Cars & Food & MNIST & OxfordPet & Flowers & SUN397 & Aircraft & Caltech101 & DTD & EuroSAT & CIFAR100 & Mean \\ \hline
        $\bm{\mathit{Last}}\uparrow$ & ~ & ~ & ~ & ~ & ~ & ~ & ~ & ~ & ~ & ~ & ~ & ~ \\ \hline
        CODA-Prompt & 83.22  & 90.12  & 97.55  & 93.32  & 83.93  & 75.52  & \textbf{50.35}  & 93.55  & \textbf{78.19}  & 88.22  & 83.76  & 83.43  \\ 
        LAE & \textbf{86.00}  & 90.01  & \textbf{98.01}  & 91.66  & 78.16  & 75.04  & 39.99  & 91.47  & 76.12  & 75.67  & 85.18  & 80.66  \\ 
        CLIP & 64.56  & 88.70  & 59.43  & 89.13  & 70.53  & 65.45  & 24.42  & 87.79  & 45.11  & 54.65  & 67.36  & 65.19  \\ 
        ZSCL & 77.90  & 91.14  & 97.68  & 93.08  & 86.47  & 78.17  & 45.60  & 92.28  & 72.45  & \textbf{96.11}  & \textbf{86.01}  & 83.35  \\
        CoLeCLIP  & 83.31  & \textbf{91.15}  & 97.72  & \textbf{94.60}  & \textbf{96.19}  & \textbf{79.62}  & 48.66  & \textbf{94.82}  & \textbf{78.19}  & 83.72  & 79.30  & \textbf{84.30}  \\ \hline
        $\bm{\mathit{Forgetting}}\uparrow$ & ~ & ~ & ~ & ~ & ~ & ~ & ~ & ~ & ~ & ~ & ~ & ~ \\ \hline
        CODA-Prompt & 83.25  & 90.07  & 98.41  & 93.14  & 91.55  & 78.07  & \textbf{49.87}  & 94.51  & 77.84  & 92.56  & 83.76  & 84.82  \\ 
        LAE & \textbf{86.63}  & 90.27  & \textbf{98.60}  & 92.51  & 83.03  & 77.86  & 42.00  & 92.81  & 77.29  & 86.31  & 85.18  & 82.95  \\ 
        CLIP & 64.56  & 88.70  & 59.43  & 89.13  & 70.53  & 65.45  & 24.42  & 87.79  & 45.11  & 54.65  & 67.36  & 65.19  \\ 
        ZSCL & 81.56  & \textbf{91.57}  & 98.59  & 93.63  & 90.43  & 78.62  & 49.02  & 92.34  & 74.34  & \textbf{96.98}  & \textbf{86.01}  & \textbf{84.83}  \\ 
        CoLeCLIP  & 83.31  & 91.19  & 97.72  & \textbf{94.60}  & \textbf{96.24}  & \textbf{79.70}  & 48.66  & \textbf{94.82}  & \textbf{78.19}  & 88.49  & 79.30  & 84.75  \\ \hline
        $\bm{\mathit{Avg}}\uparrow$ & ~ & ~ & ~ & ~ & ~ & ~ & ~ & ~ & ~ & ~ & ~ & ~ \\ \hline
        CLIP & 64.56  & 88.70  & 59.43  & 89.13  & 70.53  & 65.45  & 24.42  & 87.79  & 45.11  & 54.65  & 67.36  & 65.19  \\ 
        ZSCL & 81.56  & \textbf{91.29}  & \textbf{91.17}  & 91.29  & 81.77  & 72.41  & 33.75  & 89.74  & 53.44  & \textbf{62.36}  & \textbf{69.44}  & 74.38  \\ 
        CoLeCLIP  & \textbf{83.31}  & 90.96  & 90.76  & \textbf{93.11}  & \textbf{86.89}  & \textbf{73.22}  & \textbf{35.44}  & \textbf{89.84}  & \textbf{54.13}  & 60.28  & 62.52  & \textbf{74.59} \\ \hline
    \end{tabular}
    }
     \caption{$\mathit{Avg}$, $\mathit{Last}$, and $\mathit{Forgetting}$ results (\%) in \textbf{ODCL-TIL} for Order II. CODA-Prompt and LAE are not applicable to $\mathit{Avg}$.}
    \label{table:TIL_order2}
\end{table*}

\begin{table*}[!ht]
    \centering
    \resizebox{0.8\linewidth}{!}{
    \begin{tabular}{c c c c c c c c c c c c}
    \hline
        Method & Food & MNIST & OxfordPet & Flowers & SUN397 & Aircraft & Caltech101 & DTD & EuroSAT & CIFAR100 & Mean \\ \hline
        CLIP & \textbf{88.70}  & \textbf{59.43}  & \textbf{89.13}  & \textbf{70.53}  & \textbf{65.45}  & \textbf{24.42}  & 87.79  & 45.11  & 54.65  & 67.36  & \textbf{65.26}  \\ 
        ZSCL & 88.54  & 57.76  & 85.05  & 66.62  & 64.97  & 21.03  & \textbf{88.25}  & \textbf{45.60}  & \textbf{54.67}  & \textbf{67.78}  & 64.03  \\ 
        CoLeCLIP & \textbf{88.70}  & \textbf{59.43}  & \textbf{89.13}  & \textbf{70.53}  & \textbf{65.45}  & \textbf{24.42}  & 87.00  & 45.11  & 54.01  & 60.85  & 64.46 \\ \hline
    \end{tabular}
    }
     \caption{$\mathit{Transfer}$ results (\%) of CoLeCLIP and its two competing methods that can work on unseen domains for Order II. Note that the results are applicable to ODCL-TIL only as we need to know the task IDs to justify whether they are seen or unseen domains.
    }
    \label{table:t_order2}
\end{table*}

\noindent \textbf{Performance on Both Seen and Unseen Domains.} As shown in \cref{table:CIL_order2} and \cref{table:TIL_order2}, CoLeCLIP achieves the best $\mathit{Avg}$ performance under both ODCL-CIL and ODCL-TIL settings, consistent with the findings in Order I.

\noindent \textbf{Generalization to Unseen Domains.} As shown in Tab. \ref{table:t_order2}, compared with ZSCL, which shows a 1.23\% decline in the $\mathit{Transfer}$ metric, CoLeCLIP demonstrates less forgetting of zero-shot knowledge, with a decrease of only 0.8\%.